\documentclass[10pt,twocolumn,letterpaper]{article}

\usepackage{wacv}              

\usepackage{graphicx}
\usepackage{amsmath}
\usepackage{amssymb}
\usepackage{booktabs}
\usepackage[accsupp]{axessibility}

\usepackage[pagebackref,breaklinks,colorlinks]{hyperref}

\usepackage[capitalize]{cleveref}
\crefname{section}{Sec.}{Secs.}
\Crefname{section}{Section}{Sections}
\Crefname{table}{Table}{Tables}
\crefname{table}{Tab.}{Tabs.}

\def\wacvPaperID{1440} 
\def\confName{WACV}
\def\confYear{2024}

\begin{document}

\title{3D Face Style Transfer with a Hybrid Solution of NeRF and Mesh Rasterization}

\author{Jianwei Feng\\
Amazon\\
{\tt\small jianwef@amazon.com}
\and
Prateek Singhal\\
Amazon\\
{\tt\small prtksngh@amazon.com}
}
\twocolumn[{
\maketitle
\begin{center}
    \captionsetup{type=figure}
    \includegraphics[width=.98\textwidth]{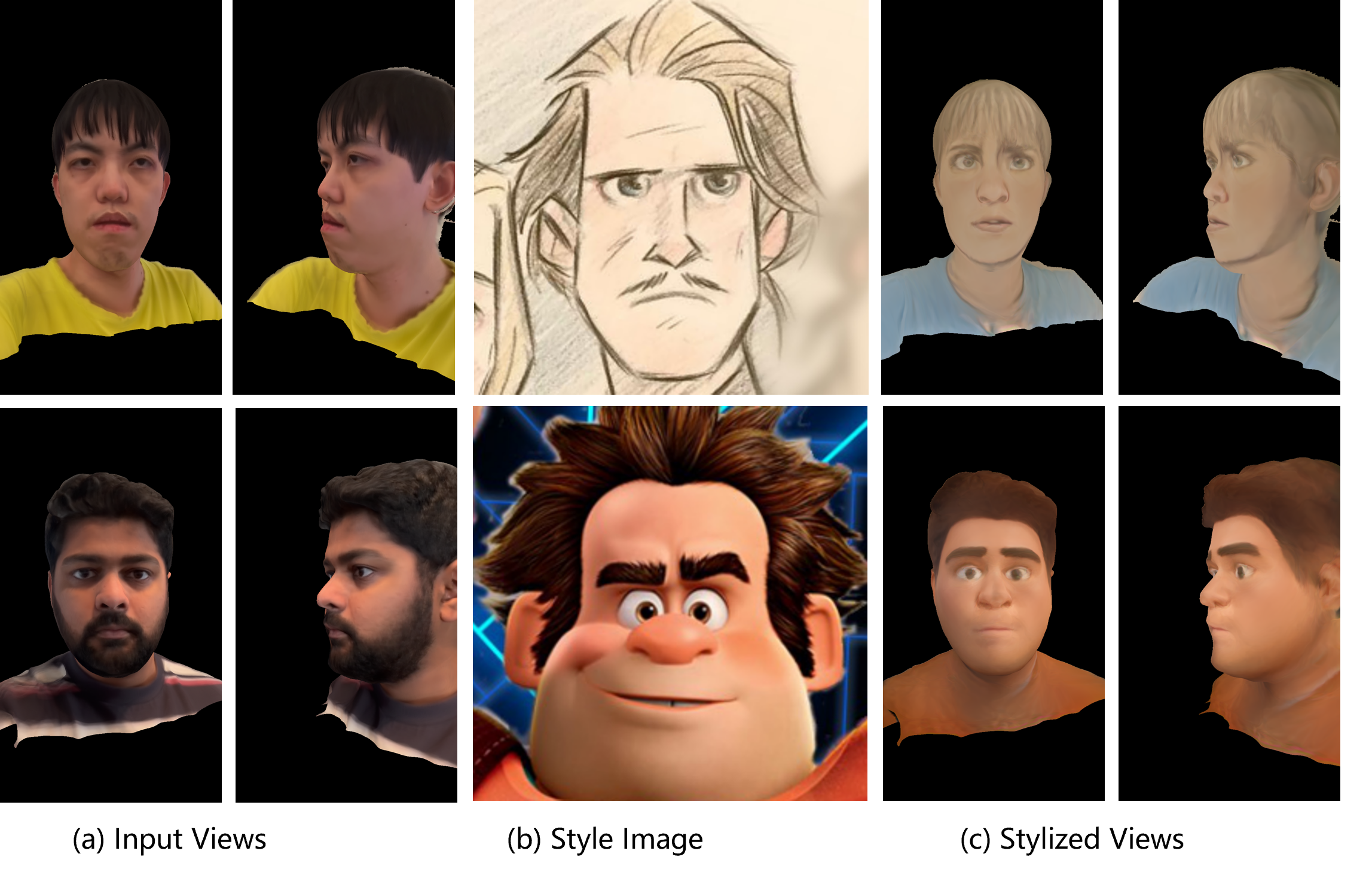}
    \captionof{figure}{Given a set of multi-view input images of a human face (a), our approach reconstructs a 3D human face, transfers the style of a style image (b) to it and generates 3D consistent stylized novel views of the face (c).}

\end{center}

}]

\thispagestyle{empty}

\begin{abstract}
Style transfer for human face has been widely researched in recent years. Majority of the existing approaches work in 2D image domain and have 3D inconsistency issue when applied on different viewpoints of the same face. In this paper, we tackle the problem of 3D face style transfer which aims at generating stylized novel views of a 3D human face with multi-view consistency. We propose to use a neural radiance field (NeRF) to represent 3D human face and combine it with 2D style transfer to stylize the 3D face. We find that directly training a NeRF on stylized images from 2D style transfer brings in 3D inconsistency issue and causes blurriness. On the other hand, training a NeRF jointly with 2D style transfer objectives shows poor convergence due to the identity and head pose gap between style image and content image. It also poses challenge in training time and memory due to the need of volume rendering for full image to apply style transfer loss functions. We therefore propose a hybrid framework of NeRF and mesh rasterization to combine the benefits of high fidelity geometry reconstruction of NeRF and fast rendering speed of mesh. Our framework consists of three stages: 1. Training a NeRF model on input face images to learn the 3D geometry; 2. Extracting a mesh from the trained NeRF model and optimizing it with style transfer objectives via differentiable rasterization; 3. Training a new color network in NeRF conditioned on a style embedding to enable arbitrary style transfer to the 3D face. Experiment results show that our approach generates high quality face style transfer with great 3D consistency, while also enabling a flexible style control.
\end{abstract}

\section{Introduction}

Style transfer for human face has been a popular research area in recent years. It has various applications in animations, advertising and gaming industry. Existing style transfer approaches for human face mainly focus on 2D image domain, where the input of the system is generally a style image and a content image, and the output is a stylized image which preserves the identity of the content image while having the style of the style image. The approaches for 2D face style transfer are usually achieved by 2D convolutional neural networks and pose 3D inconsistency issue when applied on a video or multi view images of the same face, which constraints usage of these 2D style transfer approaches in movies, animations or gaming for a consistent visual experience.

Several recent studies on 3D style transfer leverage NeRF to stylize a 3D scene. They generally supervise NeRF training with style transfer objectives applied on images rendered from NeRF, which introduces training time and memory challenge due to volume rendering on large number of pixels to form the full image needed to compute style transfer losses. Stylizing-3D-Scene \cite{chiang2022stylizing} proposed a hyper network which was conditioned on style embedding of a style image and transferred style information to the color network of NeRF. They applied style transfer losses on small image patches (32x32) to avoid issues in training time and memory. UPST-NeRF \cite{chen2022upst} also utilized a hyper network and trained on small image patches. Training with small image patches has difficulty in capturing global semantic information and leads to a loss in style transfer quality. ARF \cite{zhang2022arf} proposed a nearest neighbor-based Gram matrix loss for style transfer and deferred gradient descend to optimize on full image instead of image patch. However, deferred gradient descend significantly slows down the training process as it doesn't reduce the computation needed for volume rendering full resolution image. 



To reduce training time and memory of NeRF, recent work \cite{hong2022avatarclip, zheng2022avatar} proposed to only sample points near object surface for volume rendering. In this paper, we take it one step forward and propose to use just one single surface intersection point to render, in which case the volume rendering falls back to its simplest form and becomes equivalent to rendering a mesh extracted from NeRF. Compared to volume rendering, mesh rasterization is faster and consumes less GPU memory. We then propose a three stage approach for 3D face style transfer, where we apply different 3D representation and rendering techniques in different stages to optimize for different loss objectives in consideration of their computation needs. In the first stage, we train a NeRF model to reconstruct 3D geometry from input face images, optimized by an RGB loss applied on a batch of randomly sampled pixels through volume rendering. In the second stage, we extract a mesh from the trained NeRF model, and stylize the mesh color from a style image. The mesh color is optimized by style transfer objectives applied on full image rendered from differentiable mesh rasterization\cite{laine2020modular}. We generate 200 stylized meshes from 200 style images in a training dataset. In the third stage, we fix the geometry network weight of NeRF, and train a hyper network to predict the color network weight from a style image, to generalize for arbitrary style transfer. During each training iteration, we randomly sample a style image and its corresponding stylized mesh, and renders a full image through mesh rasterization. The hyper network is then optimized by an RGB loss between a random batch of predicted pixels from NeRF's volume rendering, and corresponding pixels from mesh rendered image. With the combination of NeRF and mesh rasterization, we are able to do 3D face style transfer at original resolution of up to 2K.

During mesh optimization, we observe that using raw style image for style transfer objectives usually leads to poor convergence due to the large difference in identity and head pose with the content images rendered at different view points. We therefore propose to generate pair data of stylized images with similar head pose and identity by applying a 2D style transfer model \cite{yang2022pastiche} on content images randomly rendered at different head pose angle. Mesh optimization with pair data shows better style transfer quality on the mesh.


To summarize, our contributions are: 
\begin{itemize}
  \item We propose a novel three stage approach which achieves arbitrary 3D face style transfer with good style transfer quality and 3D consistency.
  \item We combine NeRF and mesh rasterization to optimize for different loss objectives which enables 3D face style transfer on original image resolution of up to 2K at a reasonable training cost.
  \item We propose to generate pair data of stylized images to fill the gap of head pose and identity. Optimizing mesh colors with pair data shows better style transfer quality.
\end{itemize}

\section{Related Works}

\subsection{Novel View Synthesis}
Novel View Synthesis aims at synthesizing image at arbitrary view point from a set of source images. Traditional approaches apply explicit 3D representations to model 3D scenes, such as 3D meshes \cite{buehler2001unstructured, debevec1996modeling,waechter2014let,wood2000surface}, 3D voxels \cite{ji2017surfacenet,kutulakos1999theory,qi2016volumetric,wu20153d}, point clouds\cite{aliev2020neural,meshry2019neural,niklaus20193d,wiles2020synsin}, depth maps \cite{huang2018deepmvs,liu2015learning}. They further combine the 3D geometry defined with explicit representations with appearance representations such as colors, texutre maps, light fields or neural texture. The use of explicit 3D representations of geometry either requires supervision from ground truth 3D representation or poses strong assumption on the underlying 3D geometry.

In recent years, there has been advances in neural rendering approaches with neural radiance field (NeRF) \cite{mildenhall2021nerf,zhang2020nerf++}, where a 3D scene is represented implicitly by a multi-layer perceptron (MLP). The MLP maps the 3D coordinate and camera view direction to RGB value and density, and synthesizes a novel view via volume rendering which aggregates the colors of sampled 3D points along a ray. NeRF produces high quality novel view synthesis without the need of 3D supervision or assumption on the 3D geometry. Following works extend NeRF for faster training and inference, such as representing 3D scene with hashmap\cite{muller2022instant}, or octree\cite{liu2020neural}, followed by a reduced number of MLP layers to speed up. Other works extend NeRF to improve surface capture quality, such as NeuS \cite{wang2021neus}.
\subsection{Human Face Style Transfer}
Given a content image of human face and a reference style image, human face style transfer aims to synthesize a stylized image with the style of the style image and the structure of the content image. Traditional approaches for human face style transfer mainly focus on 2D image domain. Some works realize human face style transfer with an image-to-image translation framework, where the main idea is to learn a bi-directional mapping between the real face domain and artistic face domain \cite{shao2021spatchgan,zhao2020unpaired,nizan2020breaking}. The other line of work falls on modifying and finetuning styleGAN \cite{karras2019style}. Pinkney and Adler \cite{pinkney2020resolution} first finetuned StyleGAN on cartoon data and achieved cartoon style transfer by simply applying the latent code in original StyleGAN to finetuned cartoon StyleGAN. Kwong et al. \cite{huang2021unsupervised} further swapped the convolutional layer features between original styleGAN and a finetuned cartoon styleGAN to achieve style transfer. DualStyleGAN \cite{yang2022pastiche} modified the architecture of StyleGAN by introducing explicit extrinsic style path to have a deeper control on the style transfer. As these approaches focus on 2D image domain, they usually show 3D inconsistency issue when applied on multi view images of the same face. In contrast to 2D approaches, our approach achieves style transfer in 3D domain, with visually pleasing quality while preserving 3D consistency.


\subsection{3D Scene Style Transfer}
There have been recent works \cite{chiang2022stylizing,chen2022upst,zhang2022arf,liu2023stylerf} on 3D scene style transfer which combines style transfer and novel view synthesis and aims to synthesize novel views with style from a style image while preserving the underlying 3D structure. They mainly leverage NeRF \cite{mildenhall2021nerf} as the 3D representation for the scene. These works mainly apply on in-the-wild 3D scenes and transfer the color tone of style images. However, they cannot capture the detail style patterns and semantics as required in human face style transfer. Further, to handle the training time and memory issue from NeRF, they propose solutions that may reduce style transfer quality, or fail to generalize to unseen styles. For example, \cite{chiang2022stylizing,chen2022upst} applies style transfer losses on small image patches during training, which degrades the style transfer quality as it cannot capture global semantic information. ARF \cite{zhang2022arf} proposed deferred gradient descend to train on full resolution image, which significantly slows down training and makes learning multiple styles impossible in practice. In contrast to these works, our approach focuses on 3D human face style transfer and captures local details and semantics in style transfer. We propose a novel NeRF-mesh hybrid framework which enables fast training speed at original image resolution and achieves good style transfer quality and 3D consistency.


\section{Proposed Approach}

\subsection{Overview}

\begin{figure*}[h]

\centering
   \begin{subfigure}[b]{0.99\textwidth}
   \centering
   \includegraphics[width=\textwidth]{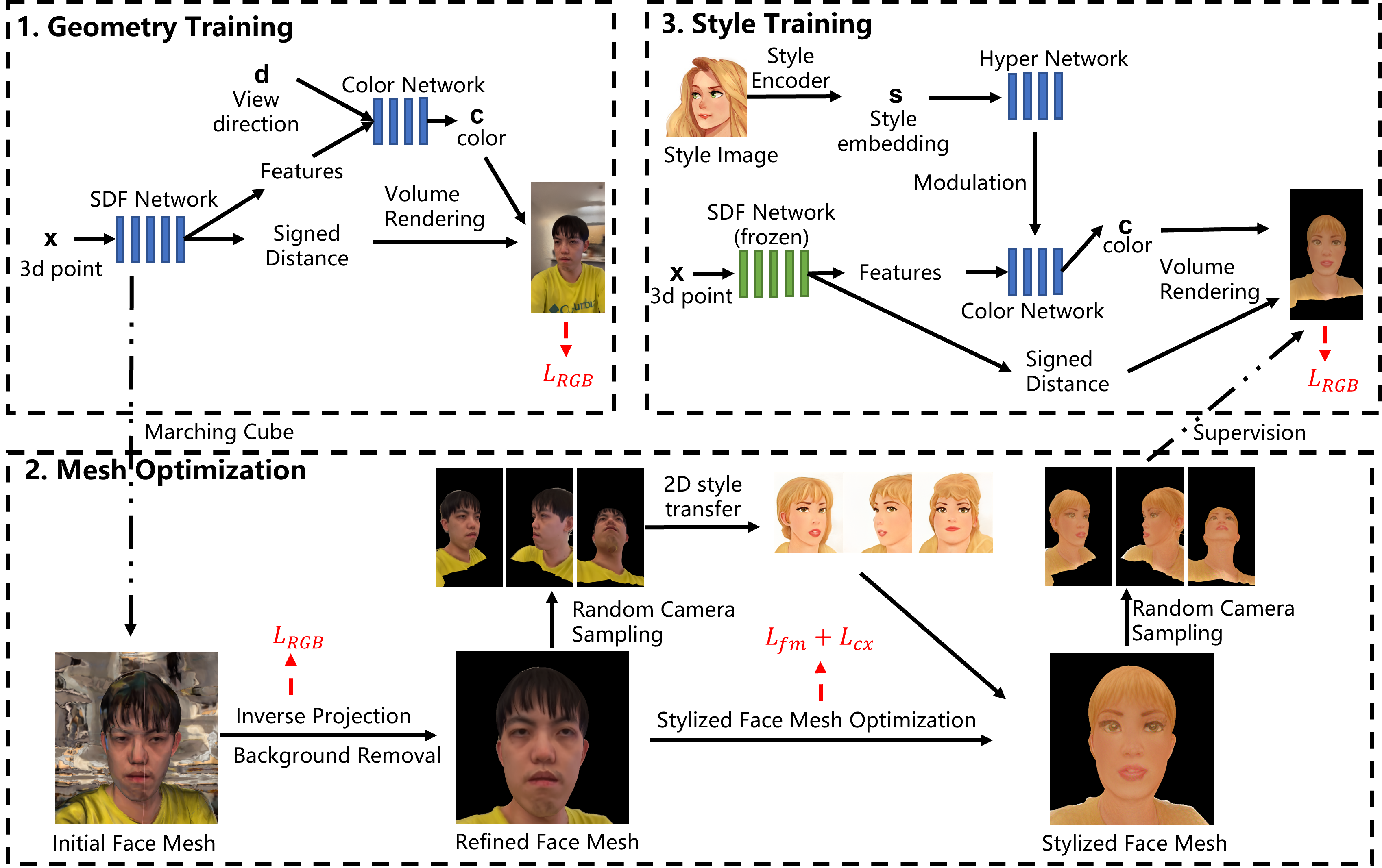}
\end{subfigure}

\caption{Overview of our approach. Our approach is in 3 stages: 1. Geometry training to learn the 3D geometry of a human face; 2. Mesh optimization to refine mesh colors and transfer style from a style image to the mesh; 3. Style training to train a hyper network conditioned on style image to generalize to arbitrary style.}
\label{figOverview}
\end{figure*}

As illustrated in Fig. \ref{figOverview}, our approach consists of three stages: 1. geometry training stage, where we train a NeRF model to capture the 3D geometry of the real face; 2. mesh optimzation stage, where we derive a mesh from the trained NeRF model, refine its color through inverse projection, and stylize it by optimizing for style transfer objectives with pair data setting; 3. style training stage, where we train a hyper network to predict NeRF's color network weight from a style embedding extracted from a style image. Details of each stage are presented in the following sections.

\begin{figure*}[h]

\centering
   \begin{subfigure}[b]{0.99\textwidth}
   \centering
   \includegraphics[width=\textwidth]{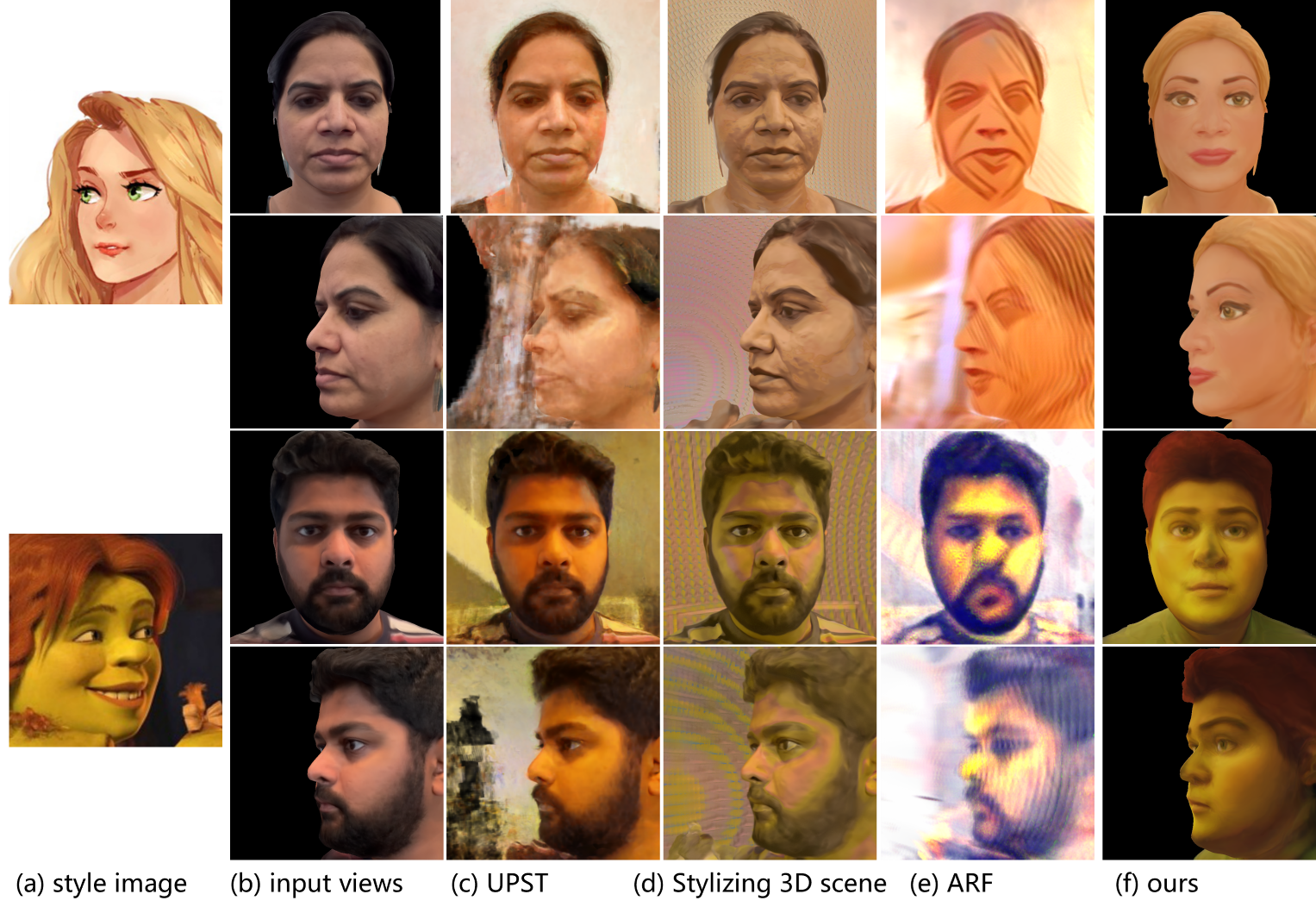}
\end{subfigure}

\caption{Qualitative Comparisons of transferring style in a style image (a) to input views (b). Our approach (f) shows better style transfer quality and 3D consistency compared to other 3D scene style transfer approaches (UPST\cite{chen2022upst} (c), Stylizing 3D Scene\cite{chiang2022stylizing} (d), ARF\cite{zhang2022arf} (e))}
\label{figQualitative}
\end{figure*}

\subsection{Geometry Training}

Neural Radience Field (NeRF) \cite{mildenhall2021nerf} uses multilayer perceptron (MLP) networks to model a 3D scene as fields of volume density and colors. Given a pixel of an image for a 3D scene at view direction, a ray from the pixel is emitted and several 3D points are sampled along the ray. For each 3D point, NeRF predicts its volume density and color by a geometry network and color network. The geometry network of NeRF maps a 3D point to volume density and features. The color network of NeRF then maps features from geometry network and view direction to RGB color. The predicted color of the pixel is derived by volume rendering which aggregates the color and volume density of the 3D points along the ray.

Original NeRF has issues with extracting high quality surface due to insufficient surface constraint during training. To derive higher quality mesh from a trained radiance field, we use NeuS \cite{wang2021neus} which proposes improvements in surface capture. NeuS represents surface as a signed distance function (SDF) and replaces geometry network in NeRF with an SDF network to predict signed distance from a 3D point. It also modifies volume rendering formulation based on SDF and introduces an extra loss terms for surface regularization.

In the geometry training stage, we train a NeuS model on input face images. The trained SDF network of NeuS represents the 3D geometry of the human face.



\subsection{Mesh Optimization}
After training a NeuS model on input images, we use marching cube \cite{lorensen1987marching} to export a face mesh from trained SDF network. To optimize face mesh, we apply differentiable rasterization \cite{laine2020modular} to render image from mesh, and apply losses on image level, where the gradients of the losses can be back propagated to the mesh. Optimizing topology of 3D mesh from image supervision usually leads to suboptimal convergence, as analyzed in \cite{liao2018deep, remelli2020meshsdf}. We therefore fix the vertex locations of the mesh and only optimize for vertex colors.

\textbf{Mesh Refinement}: The initial mesh from marching cube generally contains some artifacts in the colors. This is because the color network of NeuS model was trained with volume rendering which aggregates colors along the ray to form final color at pixel, and thus the color at surface point has some gap with the color seen in the image. We then refine the mesh color by optimizing an inverse projection problem.
\begin{equation}
    argmin_{\mathbf{c}} L_{rgb}(\mathbf{M}\odot \phi_{\mathbf{c}}(\theta), \mathbf{M}\odot \mathbf{I_{gt}})
\end{equation}

where $\mathbf{c}$ is vertex colors and $\phi_{\mathbf{c}}(\cdot)$ represents an image generator by mesh rasterization, parameterized by vertex color. $I_{gt}$ is a random ground truth image from the set of input images and $\theta$ is the corresponding view angle.  $\mathbf{M}$ is the mesh segmentation mask. We optimized the mesh color on input images with masked RGB loss with an iterative process. After inverse projection, the mesh color is refined to be similar as presented in source images. We further remove background by applying a foreground segmentation model \cite{chen2022robust} on input images and trim down mesh vertices that are visible in the input images as background pixels. After mesh refinement and background removal, the resulting mesh mainly contains human head and part of the upper body and has a photorealistic texture, which enables us to synthesize photorealistic images at different view points to use for content images for 2D style transfer. 

\textbf{Face Mesh Style Transfer}: Given a refined face mesh and a style image, we aimed at transferring the style from the style image to the face mesh through optimization. Naturally, we can view the face mesh as an image generator $\phi_{\textbf{c}}$ parameterized by vertex colors $\mathbf{c}$ that can generate content images of the face at arbitrary angle. And we apply style transfer objectives between content images and the input style image to optimize the vertex colors $\mathbf{c}$. For the style transfer objectives, we use a feature matching loss \cite{huang2017arbitrary} and contextual loss \cite{mechrez2018contextual}. This brings in our initial optimization objective below.

\begin{equation}
    argmin_{\mathbf{c}} L_{fm}(\phi_{\mathbf{c}}(\theta), \mathbf{I_{style}}) + L_{cx}(\phi_{\mathbf{c}}(\theta), \mathbf{I_{style}})
\end{equation}

where $\mathbf{I_{style}}$ is the style image, and $\theta$ is the view angle of the mesh randomly sampled from a semi sphere in each iteration of optimization.

However, the initial objectives could not optimize the mesh color to have good style transfer quality. We find that it is because of a large gap in identity and head pose between the mesh rendered images and the style image. The mesh rendered images always resemble the identity of the input images that is different with the style image. And the mesh rendered images have diverse head pose that could be largely different with style image. Therefore, we propose to optimize with pair data that has similar identity and head pose.

More specifically, instead of using a fixed style image $\mathbf{I_{style}}$ for arbitrary content image $\phi_{\mathbf{c}}(\theta)$, we use a 2D style transfer model DualStyleGAN \cite{yang2022pastiche} $\psi(\cdot)$ to generate stylized image from a content image and a style image to have similar head pose and identity with the content image. 

\begin{multline}
    argmin_{\mathbf{c}} L_{fm}(\phi_{\mathbf{c}}(\theta), \psi(\phi_{\mathbf{c}}(\theta), \mathbf{I_{style}})) + \\ L_{cx}(\phi_{\mathbf{c}}(\theta), \psi(\phi_{\mathbf{c}}(\theta), \mathbf{I_{style}})))
\end{multline}

During optimization, for each iteration, we randomly sample a view angle $\theta$ from a semi sphere, render an image $\phi_{\mathbf{c}}(\theta)$ from mesh, and generate a stylized image $\psi(\phi_{\mathbf{c}}(\theta), \mathbf{I_{style}})$ from 2D style transfer. The vertex color is optimized by the feature matching loss and contextual loss between these two.

The stylized images are generated from 2D style transfer and could contain 3D inconsistencies. As we are fixing vertex locations, the 3D consistency of the optimized mesh is guaranteed, and the optimization objectives only supervise the style of the mesh and avoid potential 3D inconsistencies from the generated stylized images. After optimization, we obtain a stylized face mesh from a style image. With mesh rasterization, the optimization is pretty fast and only takes 2 minutes per mesh.



\subsection{Style Training}
In this stage, we would like to generalize the color network of NeuS model for arbitrary style transfer. For this purpose, it should be trained with multiple styles seen so that it could generalize to unseen style. Therefore, we generate 200 stylized meshes corresponding to 200 different style images to use as our ground truth generators for training. 

We modulate the weight of the color network in NeuS model by a hyper network $\Omega(\cdot)$ whose input is a style embedding extracted from a style image by a PSP style encoder \cite{richardson2021encoding}. Given different style images, the hyper network is capable of generating different color network weight to render for different stylized outputs. 

We freeze the SDF network from stage 1 to reuse the learned 3D geometry, and only train the hyper network. We train with RGB loss supervised by the stylized mesh in stage 2. For each iteration, we randomly sample a style image $\mathbf{I_{style}}$, its corresponding stylized mesh $\phi(\cdot)$, a view angle $\theta$ at a semi sphere and a batch of pixels. We query the hyper network from a style embedding to generate weight of color network and render the color of sampled pixel through volume rendering. For RGB supervision, we use the stylized mesh $\phi(\cdot)$ to render an image from the same view angle $\theta$. Formally, 

\begin{equation}
    argmin_\Omega L_{rgb}(\Omega(\mathbf{z_{style}, \theta}), \phi(\theta))
\end{equation}
where $\mathbf{z_{style}}$ represents a style embedding from a style image, $\Omega(\mathbf{z_{style}, \theta}$ represents a batch of pixels from hyper network rendering.

In test time, the trained hyper network can be used for arbitrary style transfer. With a style image, we extract its style embedding and predict the weight of color network. And the predicted color network and the pretrained SDF network are used to generate stylized novel views through volume rendering with the style in style image applied.



\section{Experiment}

\textbf{Dataset} We collect a video dataset of 8 subjects, where each of them records a video of 10-15 seconds of 300-500 frames at 30 FPS. The videos are further processed with COLMAP\cite{schoenberger2016sfm} to estimate camera intrinsics and poses for every video frame. For style transfer, We use a cartoon dataset \cite{pinkney2020resolution} with 317 cartoon images. We use 200 images during training and hold off the remaining 117 images as unseen styles to evaluate for the generalizability of our approach. For each subject, we train a separate model for arbitrary style transfer on this subject. For a single model, the training can be finished in ~23 hours, with 7 hours for stage 1, 6 hours for stage 2 and 10 hours for stage 3.

\textbf{Methods for Comparison} We compare our approach with state of the art 3D scene style transfer approaches (UPST\cite{chen2022upst}, Stylizing 3D Scene\cite{chiang2022stylizing}, ARF\cite{zhang2022arf}), and two baselines: 1. 2D style transfer $\rightarrow$ NeuS, where we first run 2D Image style transfer \cite{yang2022pastiche} on input images and then train a NeuS model directly on top of the stylized source images; 2. Neus $\rightarrow$ 2D style transfer, where we first train a NeuS model on top of the source images to synthesize novel views for real human face, and then apply 2D style transfer on top of the synthesized novel view images.

\subsection{Qualitative Results}

We compare our approach and 3D scene style transfer approaches (UPST\cite{chen2022upst}, Stylizing 3D Scene\cite{chiang2022stylizing}, ARF\cite{zhang2022arf}) qualitatively in Fig. \ref{figQualitative}. Among the 3D scene style transfer approaches, UPST\cite{chen2022upst} is significantly under-stylized and has a bad novel view synthesis on the side view. Stylizing 3D scene\cite{chiang2022stylizing} generates 3D consistent frontal and side views, but can only transfer overall color tone and have artifacts in the background. ARF\cite{zhang2022arf} applies stronger style transfer than other two approaches, but loses details in facial structure and contains blurriness. The compared 3D scene style transfer approaches only transfer the overall color tone of the style image and fail to capture the semantics of the face, whereas our approach transfers the color of hair, skin and lip well and also achieves good 3D consistency. 


\begin{figure*}[h]

\centering
   \begin{subfigure}[b]{0.99\textwidth}
   \centering
   \includegraphics[width=\textwidth]{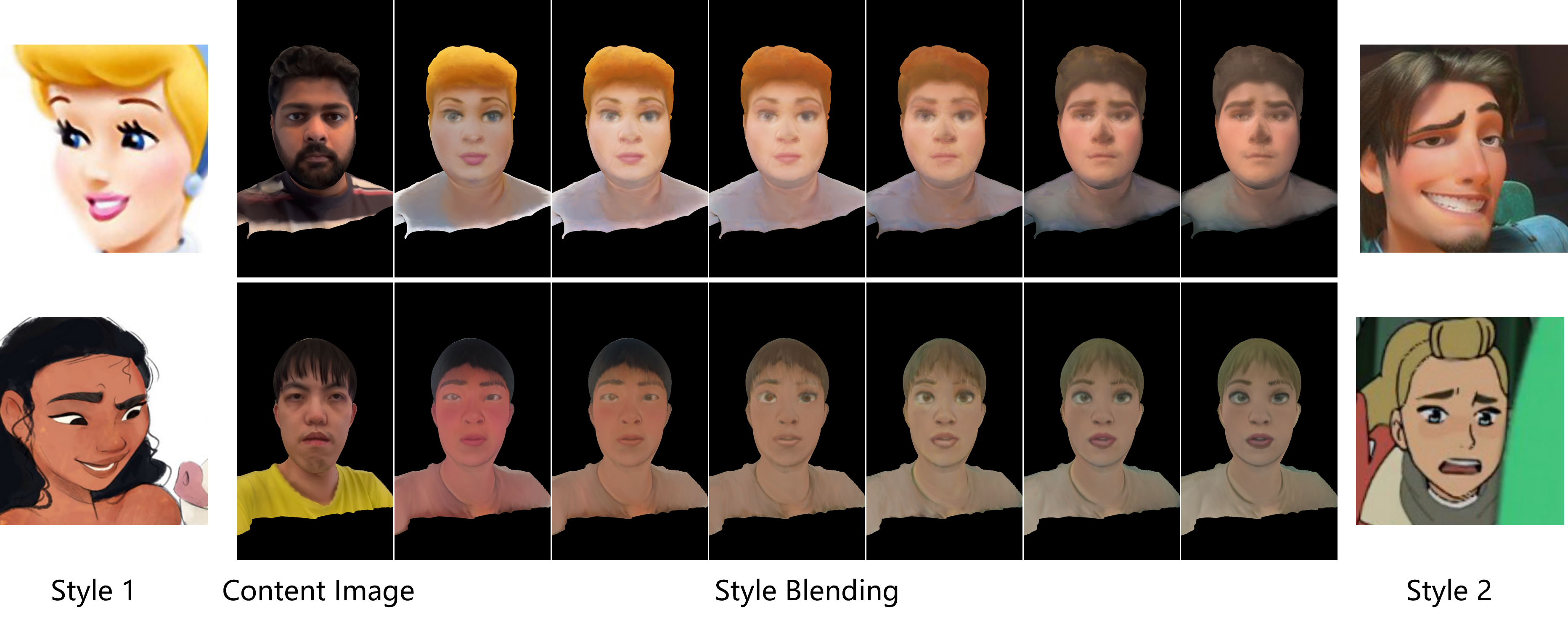}
\end{subfigure}

\caption{Style blending, our approach can interpolate between two styles and generate a mixed style of both. We show two rows of examples with style gradually changing from style 1 to style 2.}
\label{figStyleBlending}
\end{figure*}

\subsection{Quantitative Results}

\begin{table}[h!]
\caption{Quantitative Comparison on short range and long range 3D consistency error. Our approach outperforms the compared approaches by a magnitude.}
\label{TabResult3DConsistency}
\centering
\resizebox{\columnwidth}{!}{
\begin{tabular}{|c|c|c|}
\hline
Method&\begin{tabular}{c} Short Range\\ Consistency Error \\(LPIPS $\times10^{-2}$ $\downarrow$)\end{tabular} & \begin{tabular}{c} Long Range\\ Consistency Error \\(LPIPS$\times10^{-2}$ $\downarrow$) \end{tabular} \\
\hline
\begin{tabular}{c}2D style transfer \\$\rightarrow$ NeuS\end{tabular}&1.21&3.47\\
\hline
\begin{tabular}{c}NeuS $\rightarrow$\\ 2D style transfer\end{tabular}&3.23&5.07\\
\hline
\begin{tabular}{c}UPST \cite{chen2022upst}\end{tabular}&1.71&4.14\\
\hline
\begin{tabular}{c}Stylizing 3D \\Scene\cite{chiang2022stylizing}\end{tabular}&1.20&2.06\\
\hline
\begin{tabular}{c}ARF\cite{zhang2022arf}\end{tabular}&1.88&5.12\\
\hline
\begin{tabular}{c}Ours\end{tabular}&\textbf{0.29}&\textbf{0.38}\\
\hline
\end{tabular}
}
\end{table}

\textbf{Consistency Measurement} We use the short range consistency error and the long range consistency error from \cite{lai2018learning} to measure the 3D consistency between stylized images at different view points, aligned with the other 3D scene style transfer approaches. The consistency error is implemented by a warped LPIPS metric \cite{zhang2018unreasonable} where a view is warped to another view with a depth estimation. 
\begin{equation}
    E(V_i, V_j) = LPIPS(M_{ij}\odot V_i, M_{ij}\odot f_{ij}^w(V_j))
\end{equation}
where $E(V_i, V_j)$ is the consistency error between view $i$ and view $j$, $f_{ij}^w$ is the warping function and $M_{ij}$ is the warping mask. When computing LPIPS metric, only the pixels within the warping mask are taken. For short range consistency, the consistency error is computed with every adjacent frames in the testing video. For long range consistency, the consistency error is computed with all the view pairs with the gap of 7 frames.

Table. \ref{TabResult3DConsistency} shows that our approach outperforms the compared approaches by a magnitude in both short range consistency and long range consistency. The large improvement in the 3D consistency benefits from our multi stage training where explicit mesh guidance is applied. Among the other approaches, NeuS $\rightarrow$ 2D style transfer has the lowest 3D consistency as it absorbs most of the 3D consistency issues from 2D style transfer. Other NeRF based approaches show better 3D consistency but is significantly worse than our approach as they do not have explicit mesh guidance as ours which strengthen the 3D consistency.

\begin{figure}[h]

   \begin{subfigure}[b]{0.48\textwidth}
   \centering
   \includegraphics[width=\textwidth]{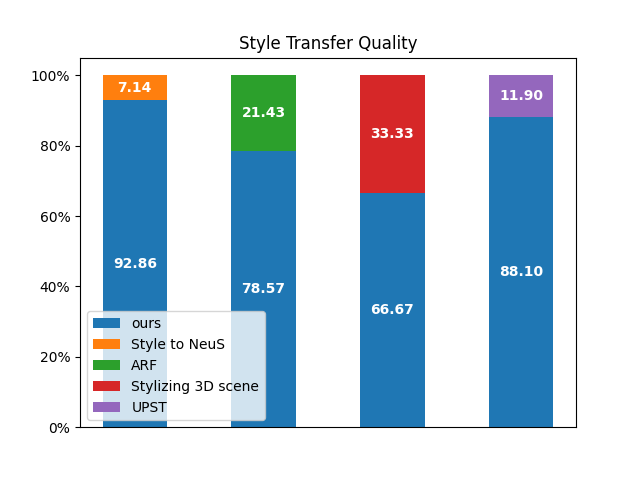}
   \caption{Style Transfer Quality}
\end{subfigure}
\begin{subfigure}[b]{0.48\textwidth}
    \centering
   \includegraphics[width=\textwidth]{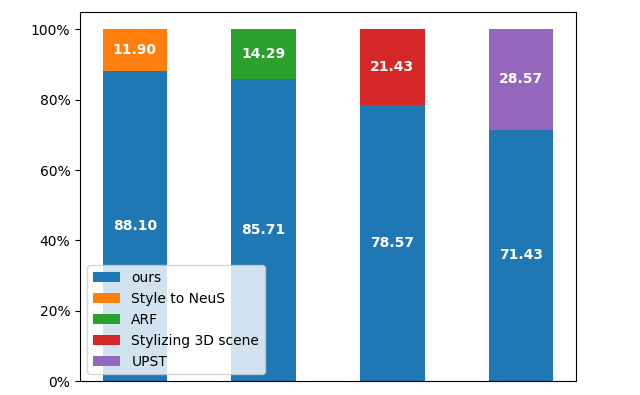}
   \caption{3D Consistency}
\end{subfigure}
\caption{User study in style transfer quality and 3D consistency. We ask the users to select the approach with better style quality or 3D consistency.}
\label{figUserStudy}
\end{figure}

\textbf{User Study} We perform a user study to evaluate the style transfer quality and 3D consistency between different approaches. We compare our approach with four different approaches (Style to NeuS, ARF\cite{zhang2022arf}, Stylizing 3D Scene\cite{chiang2022stylizing} and UPST\cite{chen2022upst}). For each comparison, we generate videos of two approaches for two identities (four videos in total). For each identity in a comparison, we ask users to make two selection: 1. select the video of better style transfer quality; 2. select the video of better 3D consistency. We collect votes from 20 participants per comparison, in total 320 votes (320=4 (comparisons) $\times$ 20 (participants) $\times$ 2 (identities) $\times$ 2 (questions). Results are shown in Fig. \ref{figUserStudy}. Our approach outperforms other approaches in both style transfer quality and 3D consistency.



\subsection{Ablation Studies}

\begin{figure}[h]
  \begin{center}
    \includegraphics[width=0.48\textwidth]{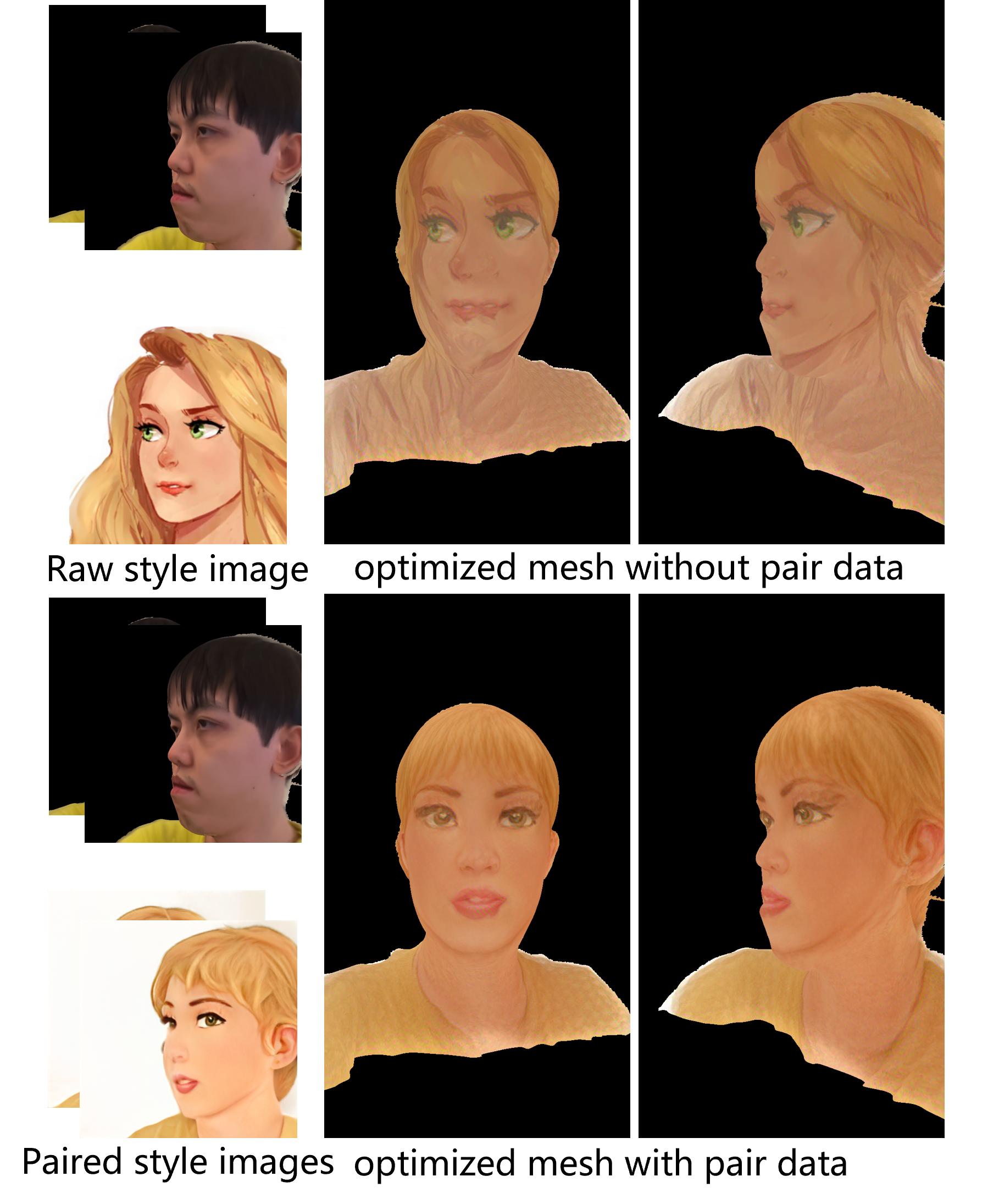}
  \end{center}
  \caption{Comparison of mesh optimization with/without pair data setting}
  \label{figAblwoPairData}
\end{figure}

\textbf{Mesh Optimization without pair data} To show the effectiveness of our pair data setting during mesh optimization stage, we do an ablation study and show that without pair data setting, the mesh optimization could not converge well, due to the large identity and head pose gap between the style image and the content image from mesh rendering. Visualization can be seen at Fig \ref{figAblwoPairData}. 

\subsection{Application}

\textbf{Style Blending} Our approach can perform smooth style blending between two styles by interpolating between the two embedding of the style images, generating smooth and harmonious style transfer of a mixed style blended from two style images, as shown in Fig. \ref{figStyleBlending}. This allows creation of non-existent styles through blending two styles.

\begin{figure}[h]
  \begin{center}
    \includegraphics[width=0.48\textwidth]{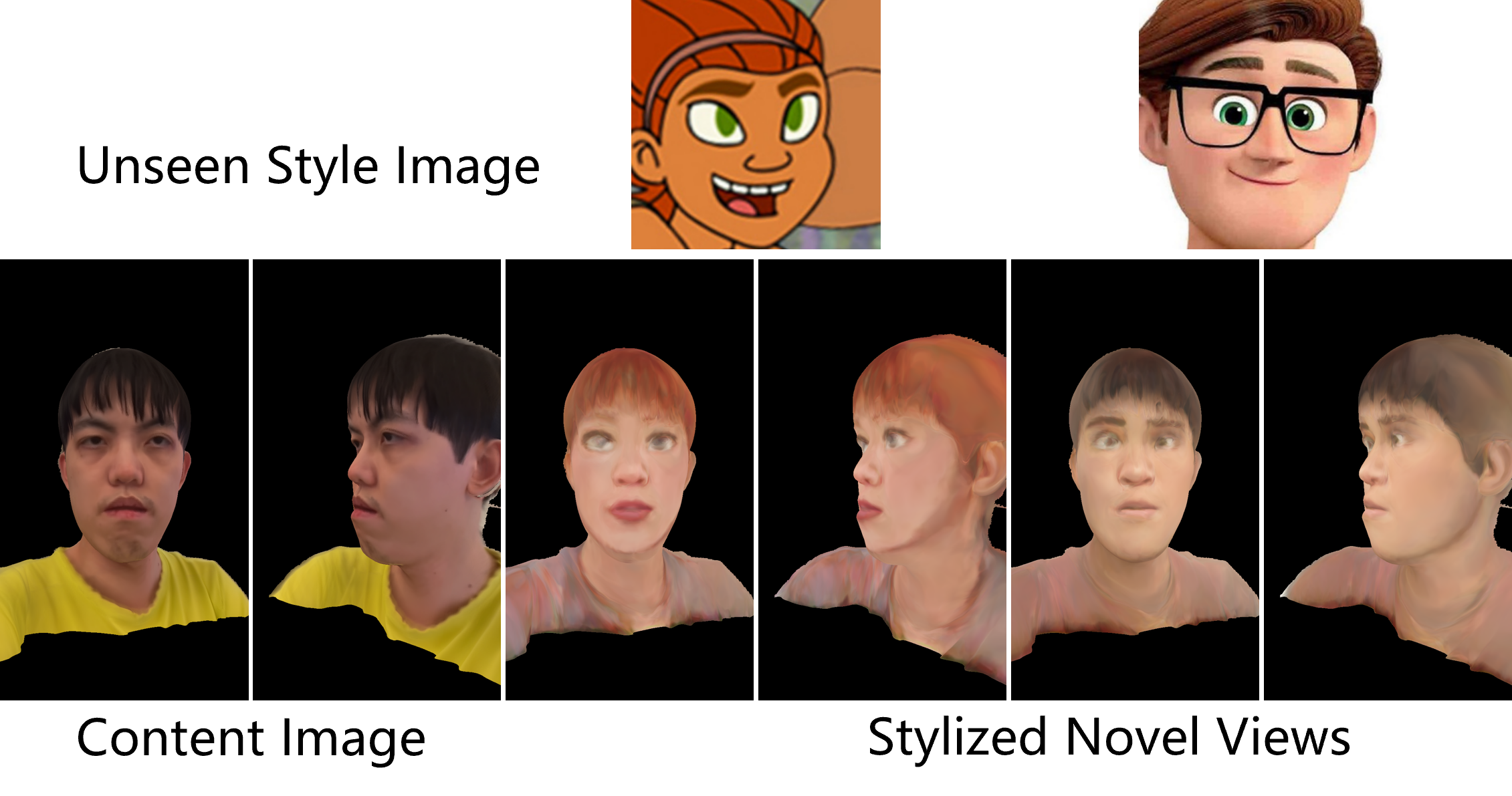}
  \end{center}
  \caption{Our approach can generalize to unseen style images and generate style transfer with decent quality and 3D consistency}
  \label{figStyleUnseen}
\end{figure}

\textbf{Unseen Style} Our approach trains a hyper network to generalize on multiple styles, hence is capable of generalizing to unseen style images in training, as illustrated in Fig. \ref{figStyleUnseen}. This allows a broader use of our approach to apply on arbitrary cartoon images for 3D human face style transfer.

\section{Conclusion}

In this paper, we propose a novel three stage approach that achieves 3D face style transfer with good style quality and 3D consistency. We present a hybrid training strategy with volume rendering and mesh rasterization which enables style transfer at original image resolution. We design a novel mesh optimization stage where we propose a pair data setting to generate decent stylized meshes. We train a hyper network on stylized meshes to generalize for arbitrary style transfer. Our experiments demonstrate that our approach outperforms baselines approaches in terms of style quality and 3D consistency quantitatively and qualitatively, and is also capable to perform smooth and harmonious style blending as well as generalizing to unseen style.

{\small
\bibliographystyle{ieee_fullname}
\bibliography{egbib}
}

\end{document}